\documentclass[10pt,twocolumn,letterpaper]{article}

\usepackage{cvpr}
\usepackage{times}
\usepackage{epsfig}
\usepackage{graphicx}
\usepackage{amsmath}
\usepackage{amssymb}
\usepackage[noblocks]{authblk}

\newcommand{\wgc}[1]{\textcolor{black}{#1}}
\usepackage[breaklinks=true,bookmarks=false]{hyperref}

\cvprfinalcopy 

\def\cvprPaperID{****} 

\ifcvprfinal\fi
\begin{document}

\title{LapEPI-Net: A Laplacian Pyramid EPI structure for Learning-based\\ Dense Light Field Reconstruction}


\author[1,2]{Gaochang Wu}
\author[1]{Yebin Liu}
\author[3]{Lu Fang}
\author[2]{Tianyou Chai}
\affil[1]{Department of Automation, Tsinghua University}
\affil[2]{State Key Laboratory of Synthetical Automation for Process Industries, Northeastern University}
\affil[3]{Tsinghua-Berkeley Shenzhen Institute \authorcr \tt\small liuyebin@mail.tsinghua.edu.cn}

\maketitle


\begin{abstract}
\vspace{-4mm}
For dense sampled light field (LF) reconstruction problem, existing approaches focus on a depth-free framework to achieve non-Lambertian performance. However, they trap in the trade-off ``either aliasing or blurring'' problem, i.e., pre-filtering the aliasing components (caused by the angular sparsity of the input LF) always leads to a blurry result. In this paper, we intend to solve this challenge by introducing an elaborately designed epipolar plane image (EPI) structure within a learning-based framework. Specifically, we start by analytically showing that decreasing the spatial scale of an EPI shows higher efficiency in addressing the aliasing problem than simply adopting pre-filtering. Accordingly, we design a Laplacian Pyramid EPI (LapEPI) structure that contains both low spatial scale EPI (for aliasing) and high-frequency residuals (for blurring) to solve the trade-off problem. We then propose a novel network architecture for the LapEPI structure, termed as LapEPI-net. To ensure the non-Lambertian performance, we adopt a transfer-learning strategy by first pre-training the network with natural images then fine-tuning it with unstructured LFs. Extensive experiments demonstrate the high performance and robustness of the proposed approach for tackling the aliasing-or-blurring problem as well as the non-Lambertian reconstruction.
\end{abstract}
\vspace{-6mm}

\section{Introduction}
\vspace{-2mm}
As an alternative to traditional 3D scene representation using geometry (or depth) and texture (or reflectance), light field (LF) achieves high-quality view synthesis without the need of such complex and heterogeneous information. More importantly, benefit from the LF rendering technology~\cite{LFrendering}, LF is capable of producing photorealistic views in real-time, regardless of the scene complexity or non-Lambertian challenges (such as jewelry, fur, glass and face, etc.)~\cite{GuoYKLY16,LFrig}. This high quality rendering requires LFs with disparities between adjacent views to be less than one pixel, i.e., the so-called densely-sampled light field (DSLF). However, typical DSLF capture either suffers from a long period of acquisition time (e.g., DSLF gantry system~\cite{LFrendering}) or falls into the well-known resolution trade-off problem, i.e., the LFs are sampled sparsely in either the angular or the spatial domain due to the limitation of the sensor resolution (e.g., Lytro~\cite{Lytro} and RayTrix~\cite{RayTrix}).

\begin{figure}
\begin{center}
\includegraphics[width=1\linewidth]{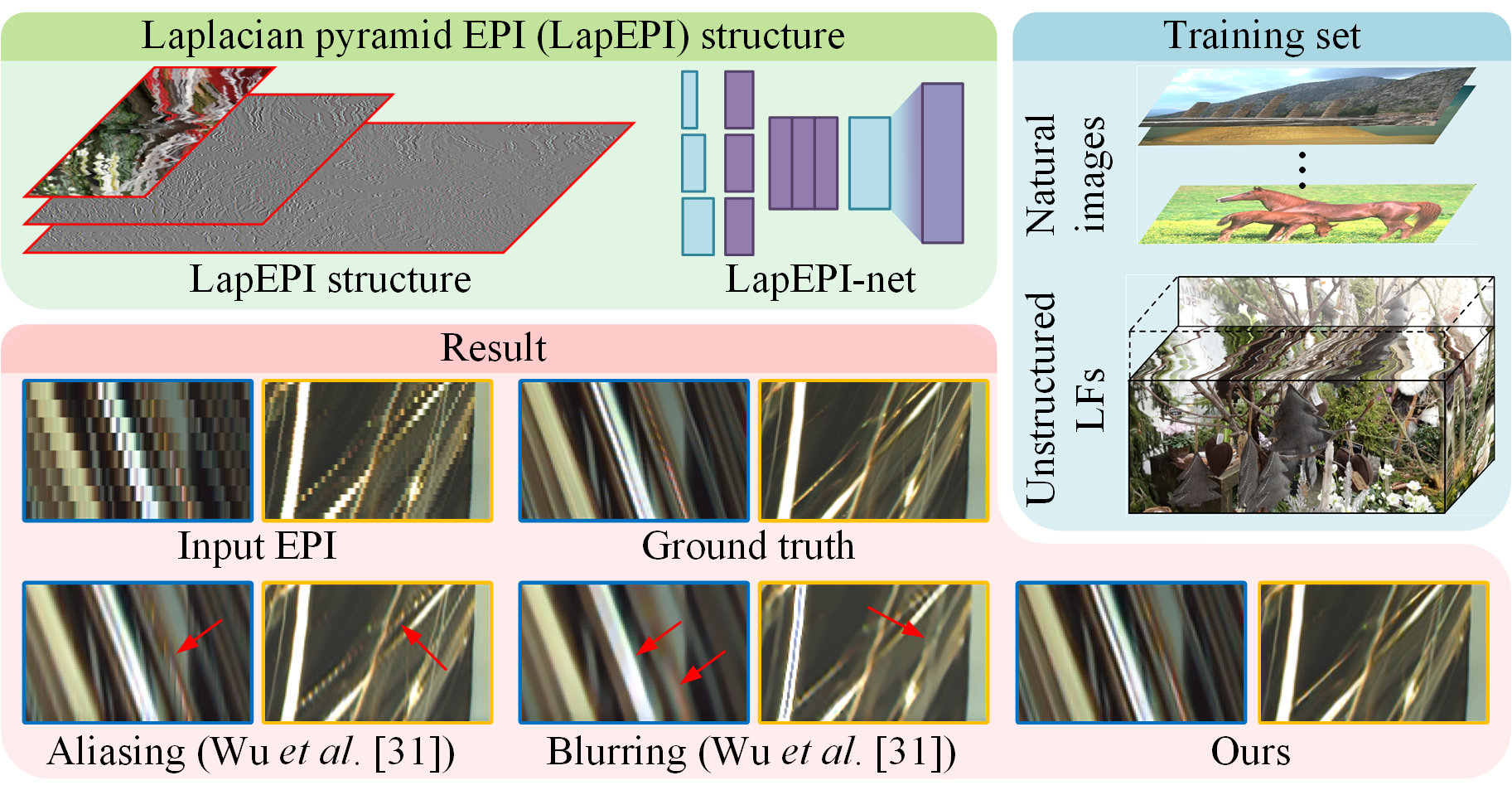}
\end{center}
\vspace{-3mm}
   \caption{We introduce a Laplacian Pyramid EPI (LapEPI) structure for solving the aliasing-or-blurring problem (pleas zoom-in to see the comparison against the results by Wu \etal~\cite{WuEPICNN2018}). The input EPI is interpolated by nearest sampling for better comparison.}
\label{fig:Teaser}
\vspace{-4mm}
\end{figure}

Recently, a more promising way is the fast capturing of a sparsely-sampled (angular domain) LF~\cite{PiCam,LFrig} followed by DSLF reconstruction~\cite{DoubleCNN,Shearlet,WuEPICNN2018}. Overall, there are two technical challenges for such DSLF reconstruction problem: angular sparsity (large disparity) and non-Lambertian. Most of the existing DSLF reconstruction methods always focus on one of the challenges but fail to solve the other. On the one hand, some approaches address the angular sparsity problem through depth estimation, and then synthesize novel views using various methods, such as soft views rendering~\cite{soft3D}, or learning-based prediction~\cite{DeepStereo}. However, the depth estimation usually based on the Lambertian assumption, namely, often fail to reconstruct the DSLF of a non-Lambertian scene. On the other hand, although approaches~\cite{LFCNN,WuEPICNN2018} circumventing depth information are able to reconstruct non-Lambertian cases, they suffer from either aliasing or blurring problem. For example, Yeung \etal~\cite{YeungECCV2018} introduced an encoder like view refinement network for larger receptive field in the angular dimensions. However, without explicitly aliasing handling, their results show aliasing effects when dealing with angular sparsity, as shown in Fig. \ref{fig:Result1}. To prevent the aliasing issue, Wu \etal~\cite{WuEPICNN2018} applied a ``blur'' operation in the spatial dimension of the input epipolar plane image (EPI), which can be considered as a pre-filter~\cite{DBLP}. But in the meanwhile, when addressing larger disparities, they had to increase the kernel size of the pre-filter accordingly, leading to a severely blurry result in the reconstructed DSLF (as shown in Fig. \ref{fig:Teaser}, bottom).



In this paper, we present a deep learning framework based on an elaborately designed EPI structure to overcome the aforementioned aliasing-or-blurring problem. Specifically, we first analyze how the variation of the spatial scale of an EPI will affect the aliasing-or-blurring performance in the Fourier domain. The analysis suggests that decreasing the spatial scale of the EPI is more efficient than simply increasing the kernel size of the pre-filter for solving the aliasing-or-blurring problem (Sec. \ref{Sec:Problem}). Then, a Laplacian Pyramid EPI (LapEPI) structure is designed to solve the aliasing-or-blurring problem by decomposing an EPI into a set of layers with different spatial scales while keeping the angular scales in the original state (Fig. \ref{fig:Teaser}, top left). Based on the LapEPI structure, we propose a novel network architecture, termed as LapEPI-net, for end-to-end EPI reconstruction. In additional, we adopt an explicit training procedure to handle the non-Lambertian challenge, i.e., a transfer-learning strategy with two accessible and efficient training sources: natural images and unstructured LFs (Fig. \ref{fig:Teaser}, right). Specifically, the network is first pre-trained on natural images then fine-tuned on unstructured LFs.

We demonstrate the competitive performance and robustness of the training sources in our LapEPI-net by performing extensive evaluations on various LF datasets. In summary, we make the following technical contributions\footnote{The code of this work will be available at:\\ \url{https://github.com/GaochangWu/LapEPI.git}}:

\vspace{-1mm}
\begin{itemize}
    \item We present a Fourier analysis of the non-linear relation between the aliasing-or-blurring problem and the spatial scales of an EPI, and design a Laplacian pyramid EPI (LapEPI) structure accordingly. The structure decomposes the input EPI into a low-spatial scale version and the residual parts with different spatial resolutions, taking the analysis conclusion as well as high-frequency components into considerations.
\vspace{-2mm}
    \item We introduce a novel network (LapEPI-net) with specifically designed architecture for the LapEPI structure. Two peculiar training sets (natural images and unstructured LFs) are adopted in a transfer-learning strategy. We demonstrate the high efficacy of the training sets for our LapEPI-net.
\end{itemize}

\vspace{-4mm}
\section{Related Work}
\vspace{-2mm}
\textbf{Depth image-based view synthesis.} These kind of approaches typically first estimate the depth of a scene~\cite{Tao,Occ,schilling2018trust,alperovich2018light}, such as structure tensor-based local direction estimation in the EPI domain~\cite{Wanner}, and phase-based depth estimation in the image domain~\cite{binolf}. Then the input views are warped to novel viewpoints and blended in different manners, e.g., soft blending~\cite{soft3D} and learning-based synthesis~\cite{Zheng2018ECCV}. In recent years, some studies for maximizing the quality of synthetic views have been presented that are based on CNNs. Flynn \etal~\cite{DeepStereo} proposed a deep learning method to synthesize novel views using a sequence of images with wide baselines. Kalantari \etal~\cite{DoubleCNN} used two sequential convolutional neural networks to model depth and color estimation simultaneously by minimizing the error between synthetic views and ground truth images. Zhou \etal~\cite{zhou2018stereo} trained a network that infers alpha and multiplane images. The novel views are synthesized using homography and alpha composition. However, all these approaches are based on the Lambertian assumption without explicitly addressing the non-Lambertian challenge.



\textbf{DSLF reconstruction without depth.} Main obstacle for DSLF reconstruction without exploiting depth information is the contradiction between aliasing effect and over-blurring~\cite{wu2017light}. Some researchers considered the reconstruction as a spectrum recovery in the Fourier domain. Shi \etal~\cite{LFfourier} performed DSLF reconstruction as an optimization for sparsity in the continuous Fourier domain. Vagharshakyan \etal~\cite{Shearlet} utilized an adapted discrete shearlet transform to remove the high-frequency spectrums that introduce aliasing effects in the Fourier domain.

Recently, some learning-based approaches were also proposed for depth-free reconstruction~\cite{wang2018end}. Yeung \etal~\cite{YeungECCV2018} applied a coarse-to-fine model using an encoder like view refinement network for larger receptive field. However, the network appears aliasing effects when reconstructing light field with large disparities. For explicitly addressing the aliasing effects, Wu \emph{et al}.~\cite{WuEPICNN2018} took advantage of the clear texture structure of the EPI and proposed a ``blur-restoration-deblur'' framework. However, when applying a large blur kernel for large disparities, the approach tends to fail at recovering the high-frequency details, and thus leading to the blur effect. 

\vspace{-2mm}
\section{Pre-Analysis: Aliasing or Blurring}
\label{Sec:Problem}
\vspace{-1mm}
A 4D LF can be represented as $L(u,v,s,t)$ with two spatial dimensions $(u,v)$ and two angular dimensions $(s,t)$. Then an EPI can be acquired by gathering horizontal lines with fixed $v^*$ along a constant camera coordinate $t^*$, denoted as $E_{v^*,t^*}(u,s)$ (or $E_{u^*,s^*}(v,t)$ similarly). The extracted EPI $E_{v^*,t^*}(u,s)$ (or $E_{u^*,s^*}(v,t)$) is denoted as $E_H$ for short, where $H$ stands for high angular resolution. The low angular resolution EPI $E_L$ can be considered as a downsampled version of the high angular resolution EPI $E_H$, i.e., $E_L=E_H \downarrow$, where $\downarrow$ denotes the downsampling operation in the angular dimension ($s$ or $t$).

Taking advantage of the clear texture structure of the EPI, Wu~\etal~\cite{WuEPICNN2018} proposed a ``blur-restoration-deblur'' framework using EPIs for the DSLF reconstruction problem. The ``blur'' operation, which in practice can be considered as a pre-filtering or anti-aliasing process, extracts the low frequency components in the spatial dimension. Without introducing the depth information, the framework shows high performance on the non-Lambertian challenge. \wgc{However, the kernel size of the pre-filter is determined by the degree of aliasing (namely, the largest disparity in the input LF). The larger disparity, the bigger blur kernel, and finally leads to more high-frequency loss, even after the recovery by the ``deblur'' operation. Therefore, this framework falls into a contradiction of the trade-off between aliasing or blurring, which we call the \textit{aliasing-or-blurring} problem.}



\begin{figure}
\begin{center}
\includegraphics[width=1\linewidth]{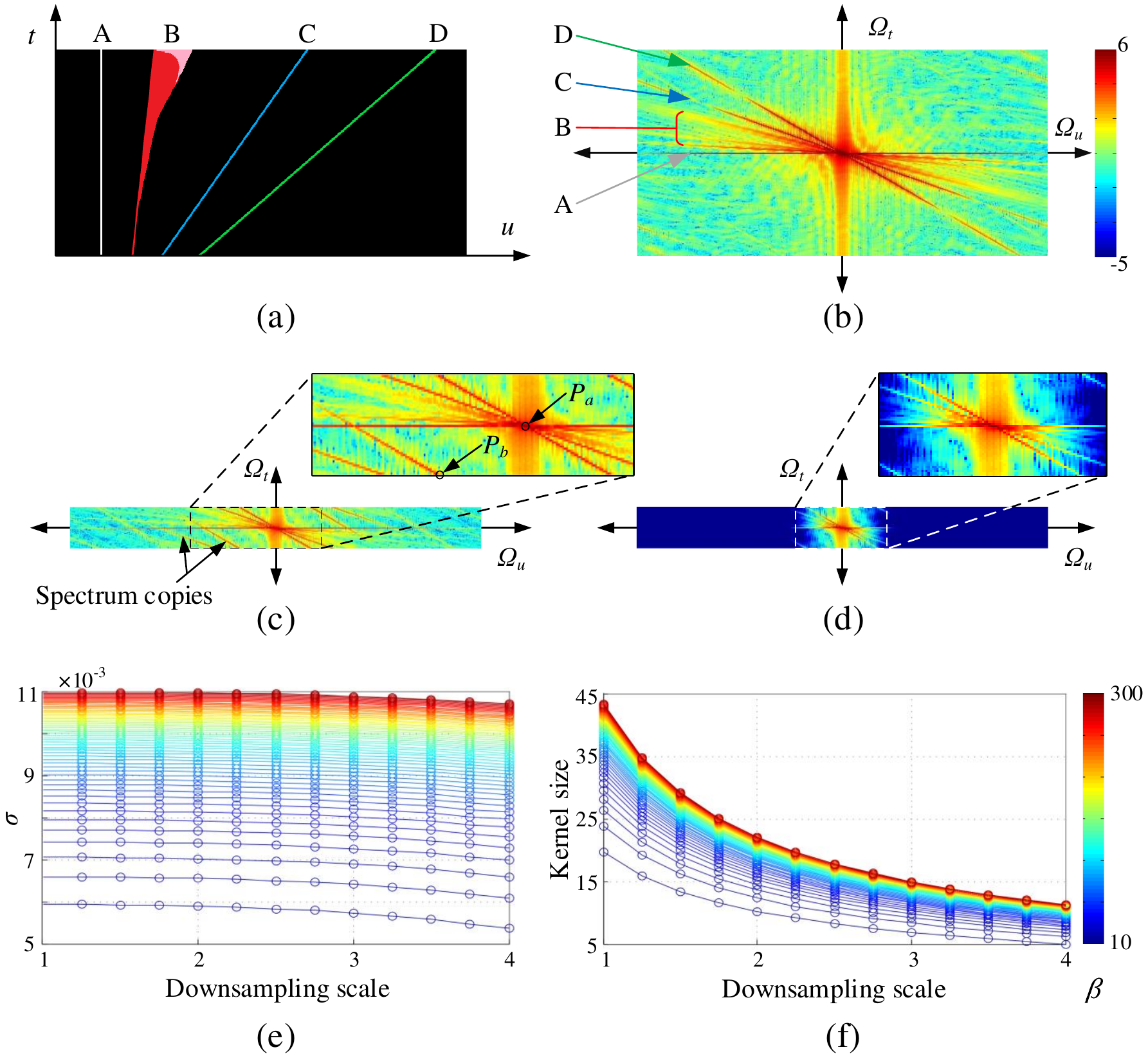}
\vspace{-7mm}
\end{center}
   \caption{Fourier analysis of the kernel size versus the downsampling scale of the input EPI $E_L$ in the spatial dimension. (a) A toy example of an EPI composed of three Lambertian objects ($A, C$ and $D$) and one non-Lambertian object ($B$) with different disparities; (b) Amplitude frequency diagram of the EPI; (c) Amplitude frequency diagram of an angularly undersampled EPI; (d) The amplitude frequency diagram of the undersampled EPI after being filtered by the designed Gaussian kernel; (e) The shape parameter $\sigma$ versus the downsampling scale and amplitude scale $\beta$; (f) The kernel size versus the downsampling scale and amplitude scale $\beta$.}
\label{fig:FA}
\vspace{-4mm}
\end{figure}

\textbf{Fourier analysis.} \wgc{Following Wu \etal~\cite{WuEPICNN2018}, we now provide a Fourier analysis of the aliasing-or-blurring problem.} Consider a simple scene composed of three points ($A, C$ and $D$) and one non-Lambertian object ($B$) located at different depths. Fig. \ref{fig:FA}(a) shows an EPI extracted from a DSLF of the scene whose disparities are within 1 pixel range between neighboring views. Fig. \ref{fig:FA}(b) plots the amplitude frequency diagram of the EPI, where the spectrum of each corresponding line (or area) in the EPI is marked with a color-coded arrow (or bracket). Note that the non-Lambertian object, $B$, appears as a set of specturms in the Fourier domain rather than a single one compared with the Lambertian points. When the EPI is angularly undersampled (i.e., the original EPI downsampled in the angular dimension), spectrum copies of an undersampled line will be produced in the high-frequency regions~\cite{WuEPICNN2018}, as shown in Fig. \ref{fig:FA}(c). The pre-filtering with kernel $\kappa$ equals to the Hadamard product between the Fourier spectrum and the Fourier transformation of the kernel. For a Gaussian blur kernel, its Fourier transformation is also a Gaussian function
\vspace{-3mm}
\begin{equation}\label{eq:Fourierkernel}
\kappa(u,\sigma)=\frac{1}{\sqrt{2\pi}\sigma}e^{-\frac{(u-c)^2}{2\sigma^2}}\Leftrightarrow\mathcal{F}_{\kappa}(\Omega_u,\sigma)=e^{-\frac{(\Omega_u-c)^2}{1/(2\pi^2\sigma^2)}},
\end{equation}
where $\mathcal{F}_{\kappa}$ denotes the the Fourier transformation of the kernel $\kappa$, $c$ is the center coordinate of the image on the $u/\Omega_u$ axis, and $\sigma$ is a shape parameter.

The kernel $\mathcal{F}_{\kappa}$ is applied to suppress the interference of the spectrum copies, namely, the anti-aliasing process. To perform a quantitative measurement, $\mathcal{F}_{\kappa}$ can be designed to reduce the amplitude of the pixel with lowest frequency among the spectrum copies (marked as point $P_b$ in Fig. \ref{fig:FA}(c)) to $1/\beta$ times of the amplitude of the center pixel (marked as point $P_a$ in Fig. \ref{fig:FA})(c)), i.e., $\mathcal{F}_{E_L}(P_b)\cdot\mathcal{F}_{\kappa}(\Omega_u(P_b),\sigma)=\mathcal{F}_{E_L}(P_a)\cdot\mathcal{F}_{\kappa}(\Omega_u(P_a),\sigma)/\beta$. Substitute into Eqn. \ref{eq:Fourierkernel}, we then have the shape parameter $\sigma$ as
\vspace{-1mm}
\begin{equation}\label{eq:shape}
\sigma=\sqrt{-\frac{\ln(|\mathcal{F}_{E_L}(P_a)|/(|\beta\cdot\mathcal{F}_{E_L}(P_b))|)}{2\pi^2\cdot(\Omega_u(P_b)-c)^2}},
\vspace{-1mm}
\end{equation}
where $\mathcal{F}_{E_L}$ is the Fourier transformation of the EPI $E_L$, and $\Omega_u(P_b)$ is the coordinate of $P_b$ on the $\Omega_u$ axis. The range of the amplitude scale parameter $\beta$ is set to $[10,300]$ in the analysis experiment. Fig. \ref{fig:FA}(d) shows the amplitude frequency diagram after being filtered by the kernel $\mathcal{F}_{\kappa}$. \wgc{When the shape parameter $\sigma$ is small, high frequencies can be preserved, but aliasing components are also remained, producing aliasing effect in the EPI. Contrarily, when the shape parameter $\sigma$ is large, aliasing components and high frequencies are removed, leading to an over-blurry result.}

\wgc{Alternatively, downsampling the input EPI $E_L$ in the spatial dimension also alleviate the aliasing effect to a certain degree. By using the quantitative measurement in Eqn. \ref{eq:shape}, we find that} when downsampling the spatial resolution of the input EPI $E_L$, the shape parameter $\sigma$ decreases non-linearly, as shown in Fig. \ref{fig:FA}(e). \wgc{In practice, we take an approximation of the infinite filter in Eqn. \ref{eq:Fourierkernel} by limiting the kernel $\kappa$ to a finite window controlled by its shape parameter $\sigma$, $u\in[-4\sigma\cdot(2c), 4\sigma\cdot(2c)]$.} We then convert the filter into the image domain with kernel size $(16\sigma c+1)$. Note that the parameter c is not a constant, but the center coordinate of the image on the $u/\Omega_u$ axis, which is inversely related to the downsampling scale. In this way, we can obtain the relation between the the kernel size and the downsampling scale, which also shows a non-linear tendency (as illustrated in Fig. \ref{fig:FA}(f)).

\textbf{Conclusion.} The analysis in Fig. \ref{fig:FA}(f) suggests that the kernel size (for the anti-aliasing processing) can be greatly reduced when downsampling the spatial resolution of the input EPI. \textit{In another words, downsampling the spatial resolution is a more effective solution for the aliasing issue, rather than simply increasing the kernel size of the pre-filter.}

\section{LapEPI Reconstruction Network}
In this section, we first introduce a novel Laplacian pyramid EPI structure according to the Fourier analysis in Sec. \ref{Sec:Problem}, which we termed LapEPI structure. We then present a novel network architecture that is specifically designed for the LapEPI structure input, termed as LapEPI-net. We also introduce how to train the LapEPI-net that is capable for reconstructing DSLFs of different scenarios from macro scenes to micro specimens. For a 3D LF $L(u,v,s)$, EPIs $E_{v^*}(u,s)$ are extracted for the reconstruction. For a 4D LF $L(u,v,s,t)$, we adopt a hierarchical reconstruction strategy described in~\cite{WuEPICNN2018}. The strategy first reconstruct EPIs using EPIs $E_{v^*,t^*}(u,s)$ and $E_{u^*,s^*}(v,t)$, then use the EPIs from the synthesized views to generate the final DSLF.

\subsection{Laplacian pyramid EPI structure}
\wgc{We have analyzed that a decreased spatial scale shows higher efficiency when handling the aliasing problem than simply increasing the kernel size of the pre-filter. Second, a multi-scale structure (different spatial scale) can be adopt for different degree of aliasing (EPIs of different disparity ranges). Third, we expect to preserve enough high-frequency components of the input EPI. Fourth, residual component in a pyramid structure removes redundant low frequency part, and thus, is more suitable and efficient for the deep leaning task, e.g., the ResNet. Combining these four elements,} we propose a Laplacian pyramid EPI (LapEPI) structure for the input EPI that contains a low spatial scale component and residual pyramid levels. Specifically, consider an input EPI $E_L$, the LapEPI structure with $P$ layers is constructed with the following steps:

1) For the first pyramid level, $\textbf{L}^1\{E_L\}=E_L\downarrow^{\alpha_s^{(P-1)}}$, where $\alpha_s$ denotes the resolution gap (upscale rate) between the neighboring levels with respect to the spatial dimension, and $\downarrow^{\alpha_s^{(P-1)}}$ represents downsampling operation in the spatial dimension with factor $\alpha_s^{(P-1)}$;

2) For the rest of the pyramid levels, compute the residual part $\textbf{R}^p(E_L)$ and convolve it with blur kernel $\kappa^p$
\begin{equation}
\begin{split}
\textbf{L}^p\{E_L\}&=\{\textbf{R}^p(E_L);\textbf{R}^p(E_L)*\kappa^p\},\\
\textbf{R}^p(E_L)&=E_L\downarrow^{\alpha_s^{P-p}}-\textbf{G}^{p-1}(E_L)\uparrow^{\alpha_s},\\
\end{split}
\vspace{-2mm}
\end{equation}
where $*$ is the convolution operation, and $\textbf{G}^p(E_L)=E_L\downarrow^{\alpha_s^{P-p}}$ denotes Gaussian pyramid in $p$\textsuperscript{th} level. To tackle the aliasing effect, we adopt pre-filters with different kernel size for the residual pyramid levels. Note that with enough spatial downsampling, the disparities is within the one-pixel range in the first pyramid level; and thus, the first level only contains the low spatial resolution image while the rest levels contain the residual part and its blurred version. We now formulate the reconstruction of EPI $E_L$ as
\vspace{-2mm}
\begin{equation}\label{Eq:objective}
f=\mathop {\min }\limits_f||E_H-f(\textbf{L}\{E_L\})||_2,
\vspace{-1mm}
\end{equation}
where $f$ is modeled as a fully-convolutional network that reconstructs the EPI in the angular dimension.

\begin{figure}
\begin{center}
\includegraphics[width=1\linewidth]{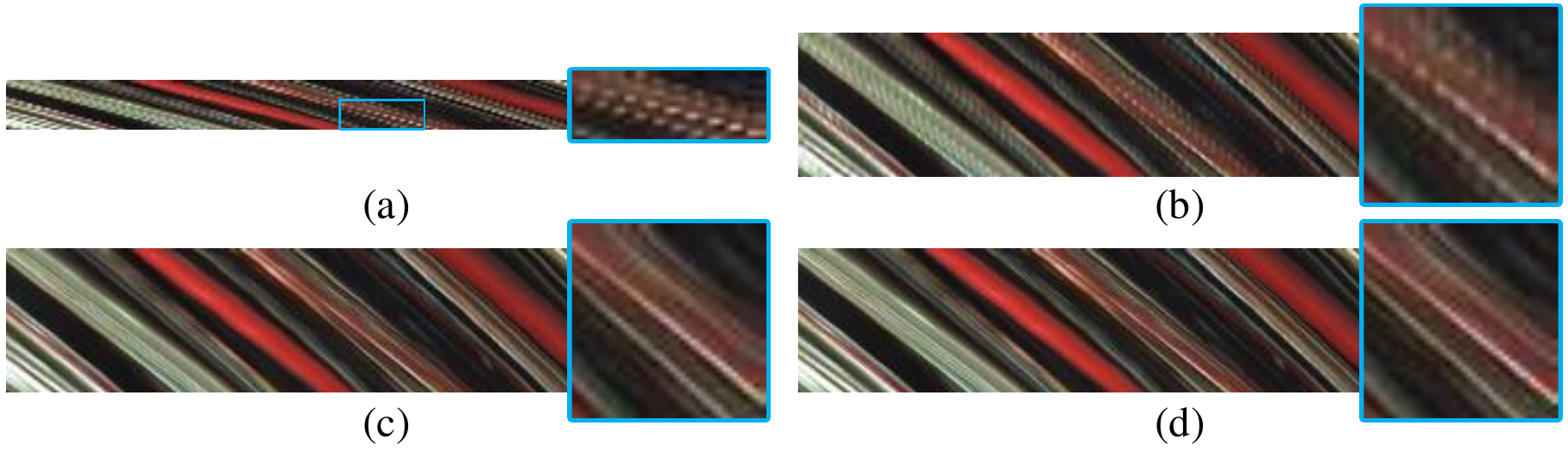}
\end{center}
\vspace{-2mm}
   \caption{Comparison of the reconstruction results with and without the proposed Laplacian pyramid EPI structure. (a) Input EPI; (b) and (c) are the results without and with the LapEPI structure under the same framework; (d) Ground truth EPI.}
\label{fig:pym}
\vspace{-4mm}
\end{figure}

We illustrate the reconstruction results with and without the proposed LapEPI structure in Fig. \ref{fig:pym}. By reducing the disparities into a small range, the proposed LapEPI structure shows a coherent and clear EPI (Fig. \ref{fig:pym}(c)) when solving the aliasing-or-blurring problem. While the result without the LapEPI structure (Fig. \ref{fig:pym}(b)) appears obvious aliasing effects in the reconstructed EPI. In this paper, the number of pyramid levels $P$ is set to 3, and the kernel size for pyramid level 2 and 3 are 5 and 13, respectively. The resolution gap between the neighboring levels $\alpha_s$ is 2.

\begin{figure*}
	\begin{center}
		\includegraphics[width=0.95\linewidth]{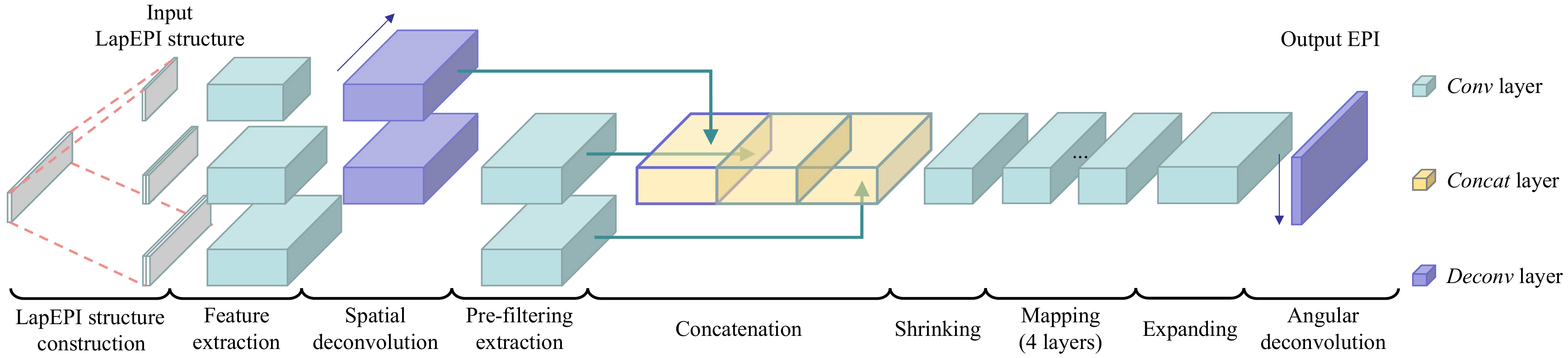}
	\end{center}
	\vspace{-3mm}
	\caption{Architecture of the proposed EPI reconstruction network (LapEPI-net). Network input is the proposed LapEPI structure, where the first level is a low spatial-angular resolution EPI and the rest of levels contain both the residual parts and their blurred versions. We use deconvolution layers to perform upsampling in both spatial and angular dimension.}
	\label{fig:CNN}
	\vspace{-3mm}
\end{figure*}

\subsection{Architecture of network}
Based on the LapEPI structure, we apply separate branches for the pyramid layers as shown in Fig. \ref{fig:CNN}. The overall network can be decomposed into eight parts: feature extraction, spatial deconvolution, pre-filtering extraction, concatenation, shrinking, mapping, expanding and angular deconvolution. The second and the last are deconvolution layers, while the rests are convolution layers. All the layers expect for the last are followed by a parametric rectified linear unit (PReLU).

The first part of layers (denoted as $Conv_{FE}^p, p={1,...,P}$) perform feature extraction separately for each pyramid level. The spatial deconvolution layers (denoted as $Deconv_{S}^p, p={1,...,P-1}$) upsample the features to the desired spatial resolution. Note that the spatial resolution of the last pyramid level equals the original, thus there is no spatial deconvolution layer being set for this level (as shown in Fig. \ref{fig:CNN}). The pre-filtering extraction layers (denoted as $Conv_{PE}^p, p={2,...,P}$) are designed to extract the blur patterns in the residual pyramids (i.e., level 2 to $P$) along the spatial dimension. We intend to set the filters in these layers with the same kernel sizes as those in the pre-filtering operation, as shown in Table \ref{table:Archi}. We then concatenate these branches together, denoted as $\{Deconv_{S}^1,Conv_{PE}^2,...,Conv_{PE}^P\}$. The branch number corresponds to the pyramid levels, i.e., $P=3$.

The following layers is an hourglass-shape structure inspired by Dong~\etal~\cite{FSRCNN}. The shrinking layer (denoted as $Conv_{S}$) is designed to reduce the feature dimension. The second part in the hourglass structure contains multiple mapping layers, denoted as $Conv_{M}^{m}$ ($m=4$ is the number of mapping layers). The last part in the hourglass structure is expanding layer (denoted as $Conv_{E}$). This layer expands the feature to a higher dimension, and can be considered as an inverse layer of the shrinking layer. The hourglass structure is able to reduce the number of parameters while keeping the high performance of the network~\cite{FSRCNN}.

The last layer in our LapEPI-net is a deconvolution layer (denoted as $Deconv_{A}$) that designed to upsample (in the angular dimension) and aggregate features from the hourglass structure. Compared with applying bicubic (or bilinear) interpolation before feeding the EPI to the network, the deconvolution layer not only reduces the size of tensors in the previous layers but also improves the performance. Table \ref{table:Archi} lists detailed parameters for each layer.

\begin{table}\small
\begin{center}
\begin{tabular}{p{3.4cm}|p{1.2cm}p{1.0cm}p{1.0cm}}
\hline
Layer & $k$ & $chn$ & $str$ \\
\hline
$Conv_{FE}^p\ (p=\{1,2,3\})$ & $5\times5$ & 56 & [1,1] \\
$Deconv_{S}^p\ \ \ \ (p=\{1,2\})$& $5\times5$ & 56 & $[1,\alpha_s^{P-p}]$ \\
$Conv_{PE}^p\ \ \ \ \ (p=\{2,3\})$& $1\times k^p$ & 56 & [1,1] \\
$Conv_{S}$ & $1\times1$ & 24 & [1,1]\\
$Conv_{M}^m(m=\{1,...,4\})$ & $3\times3$ & 24 & [1,1] \\
$Conv_{E}$ & $1\times1$ & 56 & [1,1] \\
$Deconv_{A}$ & $9\times9$ & 1 & $[\alpha_a,1]$ \\
\hline
\end{tabular}
\end{center}
\caption{Architecture of the proposed LapEPI-net, where $k=[5,13]$ corresponds to the kernel size, $chn$ is the number of channels, $str$ is the stride in the angular and the spatial dimension, and $\alpha_a$ is the upsampling scale in the angular dimension.}
\label{table:Archi}
\vspace{-2mm}
\end{table}

\subsection{Training data}
A key challenge in the training of our LapEPI-net is to obtain sufficiently large set of well structured LFs with both dense sampled in angular dimension as well as non-Lambertian property. On the one hand, existing LF datasets only provide LFs with a relatively small angular resolution, e.g., the angular resolution in HCI datasets~\cite{HCI} is 9. On the other hand, regular LFs (Lambertian scenes) lacks the information for describing non-Lambertian regions (including specular and transparent regions), which, however, are extremely common in microscopic scenes.

\begin{figure}
\begin{center}
\includegraphics[width=1\linewidth]{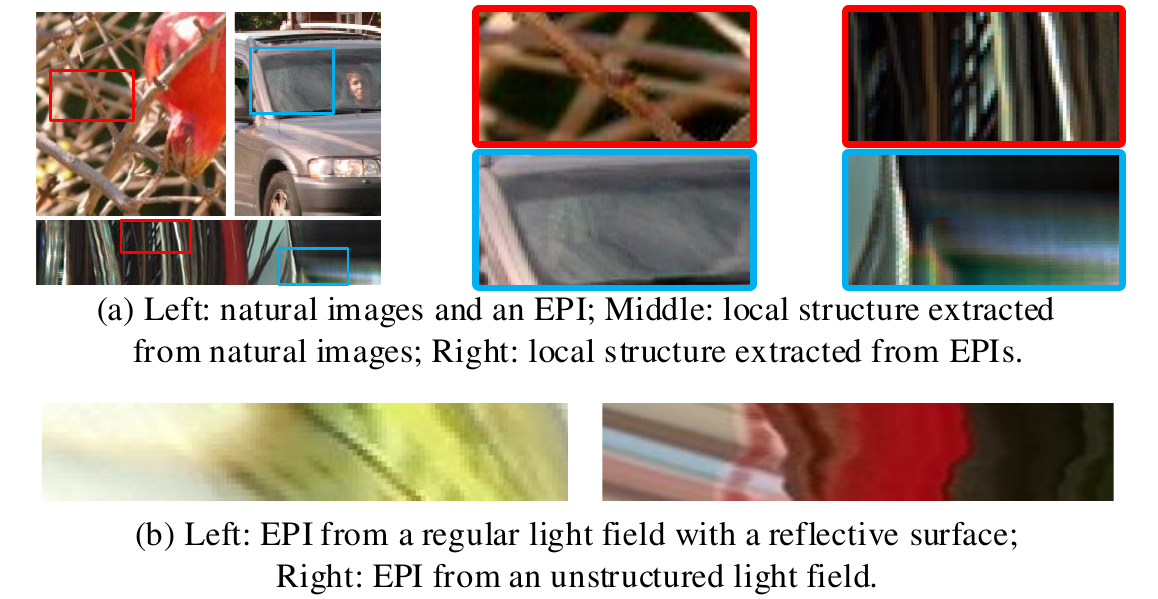}
\end{center}
\vspace{-4mm}
   \caption{An illustration of similarity in terms of local structures between natural images / unstructured LFs and regular LFs.}
\label{fig:structure}
\vspace{-2mm}
\end{figure}

Instead of concentrating on the regular LFs, we adopt two kinds of datasets for the network training: \textit{natural images} that is usually applied for single image super-resolution, and \textit{unstructured LFs} that can be acquired by a hand-hold video camera. Intuitively, from the global structure, EPIs seem quite different from natural images, thus, no one has ever attempted to extract structure prior of EPIs from natural images. However, examining them locally, the natural landscape images and EPIs share very similar local texture/structure patterns. As shown in Fig. \ref{fig:structure}(a), the visual features of a scene (such as occlusion, transparency and reflection) represented by light field appear as legible patterns in the EPI domain, e.g., straight lines, cross lines, curves, etc. The appearance property in unstructured LFs are more closer to that in regular LFs. Besides, irregular structures in an unstructured EPI also provides sufficient information similar with some particular cases in a LF, such as reflective and refractive surfaces, as shown in Fig. \ref{fig:structure}(b). More importantly, both datasets are easily collected.

\begin{figure*}
\begin{center}
\includegraphics[width=0.97\linewidth]{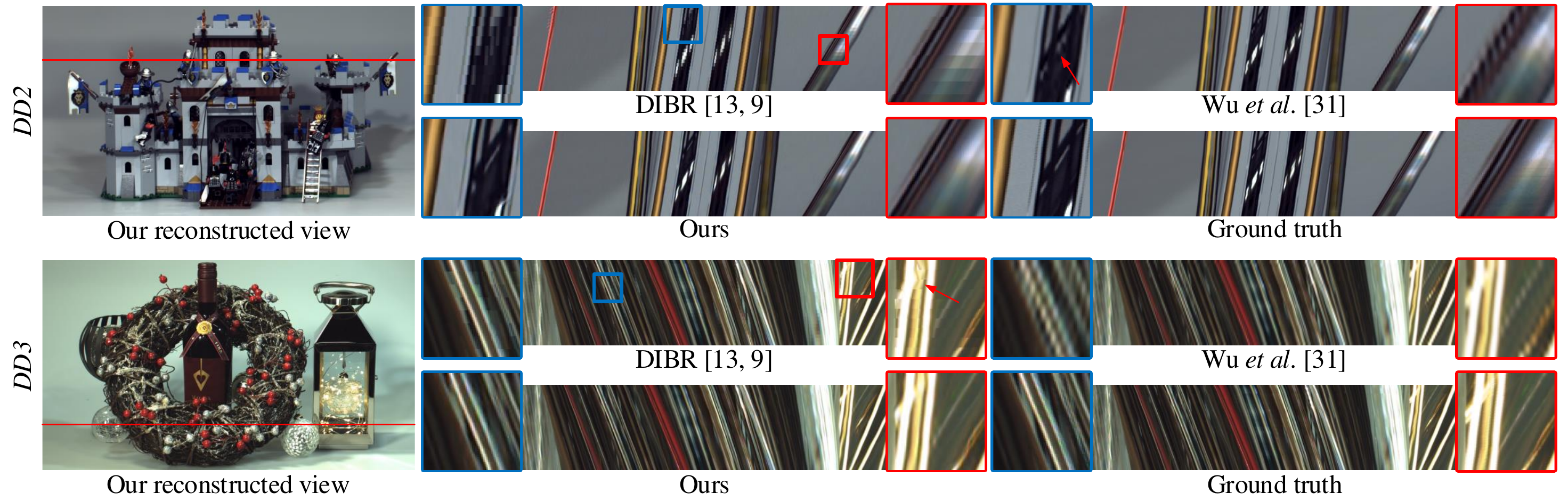}
\end{center}
\vspace{-4mm}
   \caption{Comparison of the results on the light fields from gantry system. The results show one of our reconstructed view, EPIs extracted from light fields reconstructed by each methods and the close-up versions.}
\label{fig:Result2}
\vspace{-3mm}
\end{figure*}

Specifically, we use 200 images from the BSDS200 set~\cite{BSDS200} and 91 images from Yang~\etal~\cite{T91} as the natural image dataset. In our implementation, we consider the vertical (horizontal) dimension of an natural image as the angular (spatial) dimension, and donwnsample the angular dimension using nearest sampling. For unstructured LF dataset, we use dataset provided by Y\"{u}cer~\etal~\cite{Yucer16} (including \textit{torch}, \textit{thin plant}, \textit{decoration}, \textit{plant}, \textit{scarecrow}, \textit{trunks}, \textit{basket}, \textit{orchid}, \textit{africa}, \textit{ship} and \textit{dragon}). From the unstructured LF dataset, we extract 1888 EPIs at $300\times1280$ (or 1920) (angular$\times$spatial) resolution with stride $31/400$ along dimension $v/t$.

\subsection{Implementation details}
Based on the high angular resolution EPI reconstruction objective defined in Eqn. \ref{Eq:objective}, the LapEPI-net is trained to minimize the $\mathcal{L}_2$ distance between the high angular resolution EPI label $E_H$ and the reconstructed EPI $f(\textbf{L}\{E_L\})$
\vspace{-2mm}
\begin{equation}\label{Eq:loss}
E=\frac{1}{N}\sum_{i=1}^{N}{||E_H^{(i)}-f(\textbf{L}\{E_L^{(i)}\})||_2},
\vspace{-2mm}
\end{equation}
where $N$ is the number of training samples, and $E_H^{(i)}$ and $E_L^{(i)}$ are the $i$\textsuperscript{th} high and low angular resolution EPI pair.

The upsampling scale $\alpha_a$ in the angular dimension is fixed to 3 for the training. However, the LapEPI-net is also able to reconstruct a LF with different scale, such as 2, 4, 8, etc. The training is performed on the Y channel (i.e., the luminance channel) of the YCbCr color space. To speed up the training procedure, we apply patch-wise training strategy by sampling sub-EPI pairs in the training data ($11\times44$ for the input sub-images and $31\times44$ for the labels), and mini-batches of size 28. We apply zero padding in each layers according to the filter size. Instead of training the LapEPI-net using mixed datasets from the natural images and the unstructured LFs, we divide the training into the following steps.

\textbf{Pre-training on natural images.} In this step, about 107K sub-images are extracted from 291 natural images with stride $14/20$ (dimension $y/x$). We initialize the weights of both convolution and deconvolution layers by drawing randomly from a Gaussian distribution with a zero mean and standard deviation $1\times10^{-3}$, and the biases by zero. The parameters in PReLUs are initialized as 0.1. We use ADAM solver~\cite{Kingma2014Adam} as the optimization method with learning rate of $1\times10^{-4}$ for convolution layers and $1\times10^{-5}$ for deconvolution layers, and $\beta_1=0.9$, $\beta_2=0.999$. The network converges after 600K steps of backpropagation.

\textbf{Fine-tuning on unstructured LFs.} In this step, about 705K sub-EPIs are extracted from 470 unstructured LF EPIs with stride $14/23$ (angular and spatial dimensions). The learning rate is set to $1\times10^{-5}$ for convolution layers and $1\times10^{-6}$ for deconvolution layers. The network converges after 800K steps of backpropagation. The training model is implemented using the \emph{Tensorflow} framework~\cite{TensorFlow}. The entire training steps (pre-training and fine-tuning) takes about 10 hours on a NVIDIA TITIAN Xp.

\section{Evaluations}
In this section, we evaluate the proposed LapEPI-net on several datasets, including LFs from gantry system, LFs from plenoptic camera (Lytro Illume~\cite{Lytro}) as well as microscope LFs. Note that all these datasets are not overlap with our training data in terms of data content nor LF acquisition geometry. We mainly compare our approach with three learning-based methods by Kalantari~\etal~\cite{DoubleCNN} (depth-based), Wu~\etal~\cite{WuEPICNN2018} (without depth) and Yeung~\etal~\cite{YeungECCV2018} (without depth). In addition, we perform ablation studies of our approach by training our network without pre-filtering and without using the LapEPI structure to show the performance of the proposed LapEPI structure. The quantitative evaluations is reported by measuring the average PSNR and SSIM~\cite{SSIM} values over the synthesized views. For more quantitative and qualitative evaluations, please see the submitted supplementary file and video.

\begin{table}\small
\begin{center}
\begin{tabular}{l|ccc}
\hline
& \textit{DD1} & \textit{DD2} & \textit{DD3}\\
\hline
DIBR~\cite{Acc,CDSD13}& 44.98/0.981 & 38.18/0.966 & 36.24/0.977\\
Wu~\etal~\cite{WuEPICNN2018}& 40.83/0.979 & 35.01/0.963 & 33.53/0.961\\
w/o LapEPI &41.71/0.981 & 36.95/0.970 & 34.39/0.982\\
w/o pre-filter &45.12/0.983 & 39.65/0.975 & 36.23/0.977\\
Our proposed &\textbf{46.27/0.985} & \textbf{41.96/0.978} & \textbf{40.19/0.988}\\
\hline
\end{tabular}
\end{center}
\caption{Quantitative results (PSNR/SSIM) of reconstructed LFs on the LFs from gantry system. The ``w/o LapEPI'' represent our proposed method without using the LapEPI structure.}
\label{table:Result2}
\vspace{-4mm}
\end{table}

\begin{figure*}
	\begin{center}
		\includegraphics[width=1\linewidth]{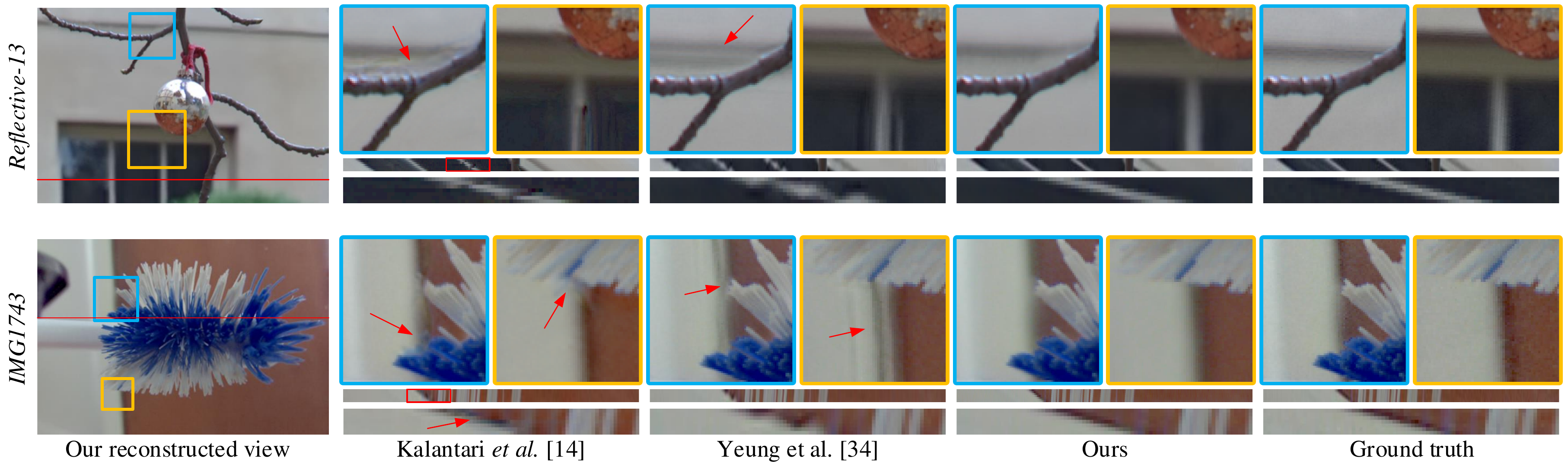}
	\end{center}
	\vspace{-4mm}
	\caption{Comparison of the results on the LFs from plenoptic camera. The results show one of our reconstructed views, close-up versions of the image portions in the yellow and blue boxes, and the EPIs located at the red line shown in the left images. We also show a close-up of the portion of the EPIs in the red box. LFs are from the Reflective category~\cite{StanfordLytro} and~\cite{DoubleCNN}.}
	\label{fig:Result1}
	\vspace{-3mm}
\end{figure*}

\subsection{LFs from gantry system}
A gantry system capture a LF by mounting a conventional camera on a mechanical gantry. Typical gantry system takes minutes to hours (depending on the angular density) to take a LF. With a high quality DSLF reconstruction / view synthesis approach, the acquisition time will be considerably reduced. For this purpose, we evaluate the proposed approach using LFs from a gantry system. The LFs are the development datasets, \textit{DD1}, \textit{DD2} and \textit{DD3}, from the ICME 2018 Grand Challenge on Densely Sampled Light Field Reconstruction~\cite{ICME2018}. In this experiment, we apply $25$ views to reconstruct a 3D LF of $193$ views. In this experiment, we apply a DIBR approach for the 3D LF view synthesis. Specifically, we use the approach by Jeon~\etal~\cite{Acc} for depth estimation and the approach by Chaurasia~\etal~\cite{CDSD13} for view synthesis. In this experiment, the performances in terms of both angular sparsity and non-Lambertian are taken into consideration.

Fig. \ref{fig:Result2} shows the results of the case \textit{DD2} and \textit{DD3}. Both cases contain challenging reflective surfaces, e.g., the ladder in the \textit{DD2} and the metal bottle in the \textit{DD3}. The depth-based view synthesis approach (DIBR~\cite{Acc,CDSD13}) outperforms the depth-free approach by Wu \etal~\cite{WuEPICNN2018}, where the latter approach shows blurring artifacts in the reconstructed EPIs. However, due to the Lambertian assumption (for both depth estimation and view synthesis) in the DIBR approach~\cite{Acc,CDSD13}, it shows sawtooth artifacts in the non-Lambertian regions. Table \ref{table:Result2} lists the quantitative measurements on these cases. In addition, two ablation evaluations of our approach are performed by training our network without pre-filtering (``w/o pre-filter'' for short) and without using the LapEPI structure (``w/o LapEPI'' for short). The numerical results demonstrate that the integrity of the proposed LapEPI structure is crucial for handling the aliasing-or-blurring problem. Our LapEPI-net takes around 2 second to reconstruct a view (spatial resolution $1280\times720$) on a 4 core CPU without GPU acceleration.

\subsection{LFs from plenoptic camera}
We evaluate the proposed approach using 32 LFs from the \textit{Refractive and Reflective surfaces} category~\cite{StanfordLytro} (Stanford Lytro Lightfield Archive) and an extra LF with large disparity from the dataset  captured by Kalantari~\etal~\cite{DoubleCNN}. In this experiment, we reconstruct a $7\times7$ LF from $3\times3$ views (extracted from the $5^\textrm{th}$ view).

\begin{table}\small
\begin{center}
\begin{tabular}{l|ccc}
\hline
& Reflective  &\textit{Reflective-13} &\textit{IMG1743}\\
\hline
Kalentary~\cite{DoubleCNN}& 37.78/0.971 & 36.01/0.963 & 40.79/0.948 \\
Wu~\etal~\cite{WuEPICNN2018}& 39.38/0.965 & 31.56/0.942 & 37.71/0.960 \\
Yeung~\etal~\cite{YeungECCV2018}& 41.35/0.968 & 33.71/0.943  & 39.07/0.948\\
Our proposed & \textbf{41.71}/ \textbf{0.976} & \textbf{40.03/0.976} & \textbf{44.57/0.980} \\
\hline
\end{tabular}
\end{center}
\vspace{-2mm}
\caption{Quantitative results (PSNR/SSIM) of reconstructed views on the LFs from plenoptic cameras. For the Reflective category, the metrics are averaged over the 32 reflective scenes.}
\label{table:Result1}
\vspace{-4mm}
\end{table}

\begin{figure*}
	\begin{center}
		\includegraphics[width=1\linewidth]{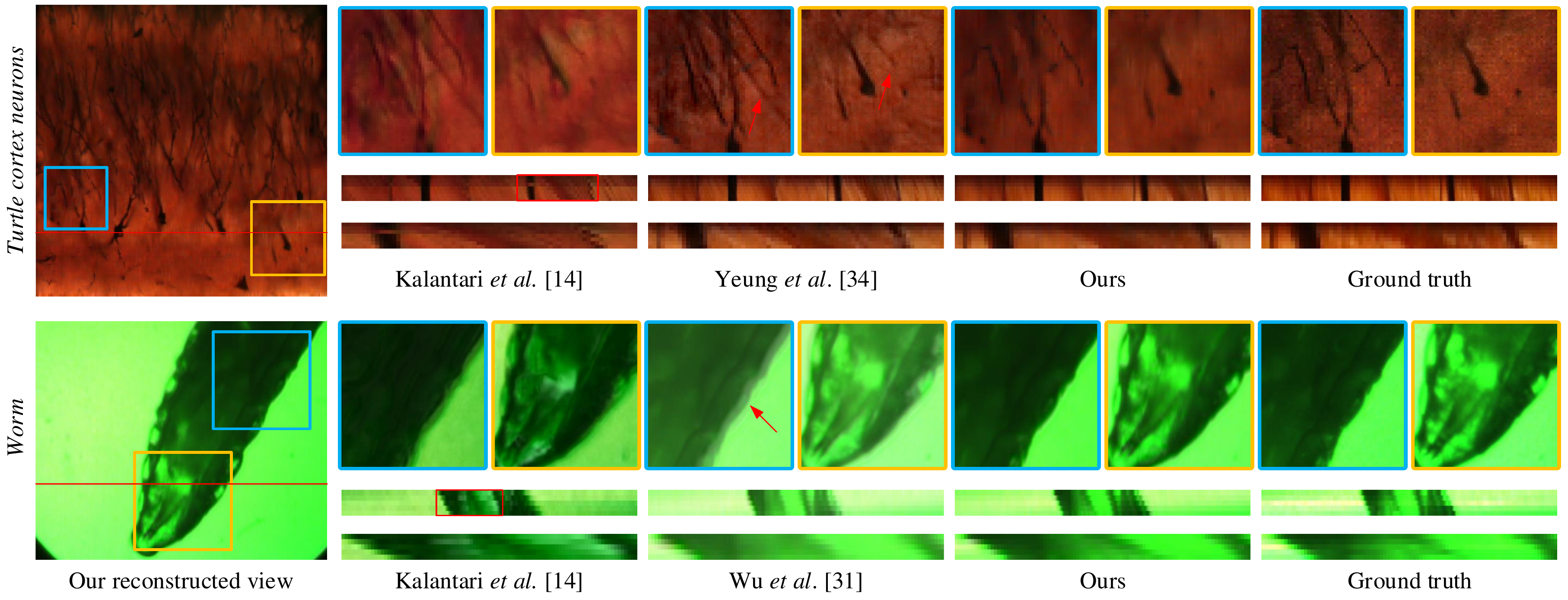}
	\end{center}
	\vspace{-4mm}
	\caption{Comparison of the results on the microscope light fields. The results show one of the synthesized results, ground truth images, a portion of the EPIs extracted from the location marked in the red line. LFs are from~\cite{microLF} and~\cite{microLFArray}.}
	\label{fig:Result3}
	\vspace{-2mm}
\end{figure*}

Fig. \ref{fig:Result1} shows the results of the case \textit{Reflective-13} from the Reflective category~\cite{StanfordLytro} and \textit{IMG1743} from the Kalantari~\etal~\cite{DoubleCNN}. The demonstrated cases contain large disparities, which conforms the aliasing-or-blurring problem. In both cases, the approach by Kalantari~\etal~\cite{DoubleCNN} produces tearing artifacts near the occlussion boundaries of the foreground objects, which can be clearly shown in the close-up images and EPIs. The approach by Yeung~\etal~\cite{YeungECCV2018} produces ghosting artifacts around the backgrounds, which are caused by the aliasing effects. In comparison, our approach provides more plausible results in terms of both local structure (shown in close-up images) and overall coherency (shown in EPIs). Table \ref{table:Result1} lists the quantitative results on the demonstrated cases and the Reflective category. The PSNR and SSIM values are averaged over the 32 reflective scenes. Although our LapEPI-net is not trained nor fine-tuned on any LFs from plenoptic camera, we show high quality results on these LFs. In terms of running time, the network takes less than 1 second to reconstruct a view on a 4 core CPU, which outperforms other two approaches~\cite{DoubleCNN,WuEPICNN2018}.

\subsection{Microscope LFs}
In this section, the Stanford Light Field Microscope Datasets~\cite{microLF} (including 2005 and 2006 datasets) and the camera array-based LF microscope datasets provided by Lin~\etal~\cite{microLFArray} (denoted as \textit{Cam. array} for short) are evaluated. We reconstruct $7\times7$ LFs from $3\times3$ (extracted from the $4^\textrm{th}$ view) views for the Stanford 2005 and 2006 datasets~\cite{microLF}, and $5\times5$ LFs from $3\times3$ views for the Cam. array datasets~\cite{microLFArray}. The approach by Yeung~\etal~\cite{YeungECCV2018} is not evaluated due to the $5\times5$ output setting.

Fig. \ref{fig:Result3} shows two representative cases with challenging occlusion relations or translucency objects. The first case (Stanford 2006~\cite{microLF}) shows a slice of Golgi-stained turtle cortex neurons with complicated occlusion relations among neurons. The LF produced by Kalantari~\etal~\cite{DoubleCNN} appears distorted artifacts, which we believe are caused by the depth errors from the depth estimation network. The approach by Yeung~\etal~\cite{YeungECCV2018} shows aliasing effect in this case around the thin structures. The second case~\cite{microLFArray} shows a translucent worm (drosophila larva) captured by a camera array system, which is challenge due to the angular sparsity as well as the non-Lambertian. The approach by Kalantari~\etal~\cite{DoubleCNN} fails to synthesize views in this transparent scenario. The result by Wu~\etal~\cite{WuEPICNN2018} shows blur effects due to the large disparities. Among these approaches, our LapEPI-net is able to provide high quality LFs in the microscope scenes. Table \ref{table:Result3} lists the quantitative results of each approach. To reconstruct a novel view, our network takes around 0.2 second for the Stanford 2005 and 2006~\cite{microLF} and 2 second for the Cam. array~\cite{microLFArray}.

\begin{table}\small
\begin{center}
\begin{tabular}{l|ccc}
\hline
&Stanford 05 &Stanford 06 & Cam. array\\
\hline
Kalantari~\cite{DoubleCNN} & 28.60/0.795 & 24.91/0.593 & 21.40/0.701 \\
Wu~\etal~\cite{WuEPICNN2018} & 29.22/0.909 & 32.36/0.885 & 27.08/0.890 \\
Yeung~\etal~\cite{YeungECCV2018} & 34.95/0.893 & 35.41/0.885 & -/- \\
Our proposed & \textbf{35.42/0.926} &\textbf{35.79/0.913} &\textbf{29.67/0.916} \\
\hline
\end{tabular}
\end{center}
\caption{Quantitative results (PSNR/SSIM) of reconstructed LFs on the microscope LFs \cite{microLF,microLFArray}. The approach by Yeung~\etal~\cite{YeungECCV2018} is not evaluated for the $5\times5$ output.}
\label{table:Result3}
\vspace{-2mm}
\end{table}


\section{Conclusions and Discussions}
Solving the aliasing-or-blurring problem is a key issue for the angular sparsity and non-Lambertian challenges. Following the existing EPI-based DSLF reconstruction framework~\cite{WuEPICNN2018}, we have first deduced a Fourier analysis of the relation between the aliasing-or-blurring problem and the variation of the spatial scale of an EPI, then designed a Laplacian pyramid EPI (LapEPI) structure. Furthermore, we have presented a novel network for the LapEPI structure, and adopted a transfer-learning strategy by first pre-training the network using natural images and then fine-tuning it with unstructured LFs. By performing evaluations on LFs from different acquisition geometries (e.g., LFs from gantry system and plenoptic camera, and microscope LFs from camera array), we have demonstrated the efficacy and robustness for solving the aliasing-or-blurring problem.

In the following, we discuss the limitations and the future work based on the presented approach. The proposed LapEPI structure have been designed to address the aliasing effect caused by the angular sparsity. With a three-pyramid-level structure, the proposed approach is capable for disparities up to 9 pixels. Increasing the number of pyramid levels may be a possible solution to enhance the capability for the angular sparsity challenge. But the network structure will have to be adjusted and retrained. \wgc{In addition, the present framework, including the LapEPI structure and the LapEPI-net, adopts fixed pre-filters. In the future, trainable pre-filters may achieve better performance.}


{\small
\bibliographystyle{ieee}
\bibliography{egbib}
}
\end{document}